\documentclass[runningheads]{llncs}
\usepackage{graphicx}
\usepackage{cite} 
\usepackage{times}
\usepackage{epsfig}
\usepackage{amsmath}
\usepackage{amssymb}
\usepackage{mwe}
\usepackage{acro}
\usepackage{amssymb}
\usepackage{xcolor,colortbl}
\usepackage{tabularx}
\usepackage{relsize}
\usepackage{pifont}
\usepackage{booktabs} 
\usepackage{multirow}
\usepackage{multicol}
\usepackage{adjustbox}
\usepackage{float}
\usepackage[colorlinks,
            linkcolor=red,  
            anchorcolor=blue, 
            citecolor=green,   
            ]{hyperref}

\begin{document}
\title{SimTriplet: Simple Triplet Representation Learning with a Single GPU}
%
%
\author{Quan Liu\inst{1} \and
Peter C. Louis \inst{2} \and
Yuzhe Lu \inst{1} \and
Aadarsh Jha \inst{1} \and
Mengyang Zhao \inst{1} \and
Ruining Deng \inst{1} \and
Tianyuan Yao \inst{1} \and
Joseph T. Roland \inst{2} \and
Haichun Yang \inst{2} \and
Shilin Zhao \inst{2} \and
Lee E. Wheless \inst{2} \and
Yuankai Huo\inst{1}}


%
\institute{
Vanderbilt University, Nashville TN 37215, USA \and
Vanderbilt University Medical Center, Nashville TN 37215, USA 
}
%
\maketitle              
\begin{abstract}
Contrastive learning is a key technique of modern self-supervised learning. The broader accessibility of earlier approaches is hindered by the need of heavy computational resources (e.g., at least 8 GPUs or 32 TPU cores), which accommodate for large-scale negative samples or momentum. The more recent SimSiam approach addresses such key limitations via stop-gradient without momentum encoders. In medical image analysis, multiple instances can be achieved from the same patient or tissue. Inspired by these advances, we propose a simple triplet representation learning (SimTriplet) approach on pathological images. The contribution of the paper is three-fold: (1) The proposed SimTriplet method takes advantage of the multi-view nature of medical images beyond self-augmentation; (2) The method maximizes both intra-sample and inter-sample similarities via triplets from positive pairs, without using negative samples; and (3) The recent mix precision training is employed to advance the training by only using a single GPU with 16GB memory. By learning from 79,000 unlabeled pathological patch images, SimTriplet achieved 10.58\% better performance compared with supervised learning. It also achieved 2.13\% better performance compared with SimSiam. Our proposed SimTriplet can achieve decent performance using only 1\% labeled data. The code and data are available at \url{https://github.com/hrlblab/SimTriplet}.

\end{abstract}

\begin{figure}[t]
\begin{center}
\includegraphics[width=0.9\linewidth]{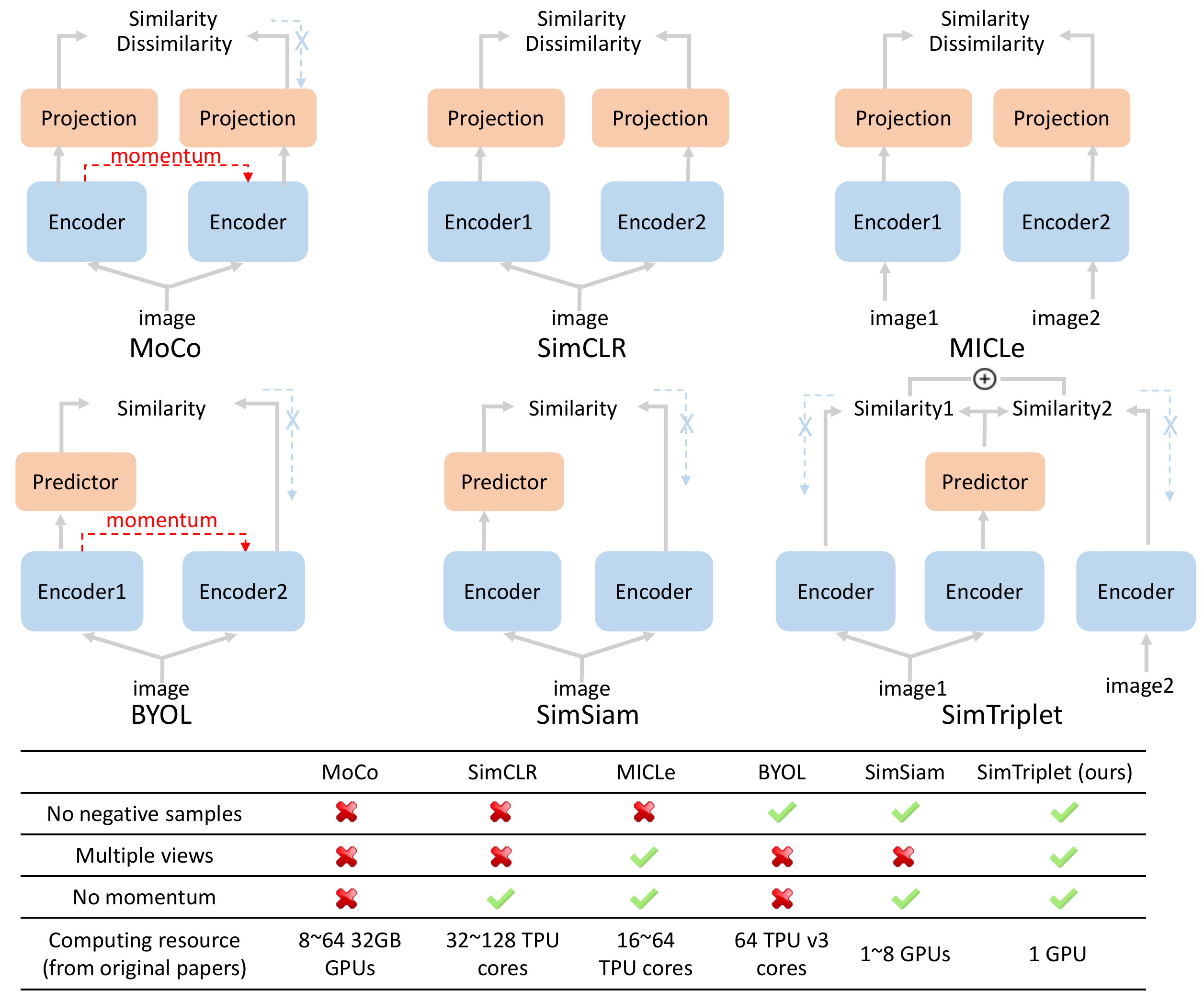}
\end{center}
   \caption{\textbf{Comparison of contrastive learning strategies.} The upper panel compares the proposed SimTriplet with current representative contrastive learning strategies. The lower panel compares different approaches via a table.}
\label{fig1:method compare}
 \end{figure}
 

\section{Introduction}
To extract clinically relevant information from GigaPixel histopathology images is essential in computer-assisted digital pathology~\cite{liskowski2016segmenting,zhu2017retinal,xu2015deep}. For instance, the Convolutional Neural Network (CNN) based method has been applied to depreciate sub-tissue types on whole slide images (WSI) so as to alleviate tedious manual efforts for pathologists~\cite{xu2017large}. However, pixel-wise annotations are resource extensive given the high resolution of the pathological images. Thus, the fully supervised learning schemes might not be scalable for large-scale studies. To minimize the need of annotation, a well-accepted learning strategy is to first learn local image features through unsupervised feature learning, and then aggregate the features with multi-instance learning or supervised learning~\cite{Hou_2016_CVPR}. 


Recently, a new family of unsupervised representation learning, called contrastive learning (Fig.~\ref{fig1:method compare}), shows its superior performance in various vision tasks ~\cite{wu2018unsupervised,noroozi2016unsupervised,zhuang2019local,hjelm2018learning}. Learning from large-scale unlabeled data, contrastive learning can learn discriminative features for downstream tasks. SimCLR~\cite{chen2020simple} maximizes the similarity between images in the same category and repels representation of different category images. Wu et al. \cite{wu2018unsupervised} uses an offline dictionary to store all data representation and randomly select training data to maximize negative pairs. MoCo~\cite{he2020momentum} introduces a momentum design to maintain a negative sample pool instead of an offline dictionary. Such works demand large batch size to include sufficient negative samples (Fig.~\ref{fig1:method compare}). To eliminate the needs of negative samples, BYOL \cite{grill2020bootstrap} was proposed to train a model with a asynchronous momentum encoder. Recently, SimSiam~\cite{chen2020exploring} was proposed to further eliminate the momentum encoder in BYOL, allowing less GPU memory consumption. 
 

To define different image patches as negative samples on pathological images is tricky since such a definition can depends on the patch size, rather than semantic differences. Therefore, it would be more proper to use nearby image patches as multi-view samples (or called positive samples) of the same tissue type~\cite{tian2019contrastive} rather than negative pairs. MICLe~\cite{azizi2021big} applied multi-view contrastive learning to medical image analysis. Note that in~\cite{tian2019contrastive,azizi2021big}, the negative pairs are still needed within the SimCLR framework.

\begin{figure}[t]
\begin{center}
\includegraphics[width=0.75\linewidth]{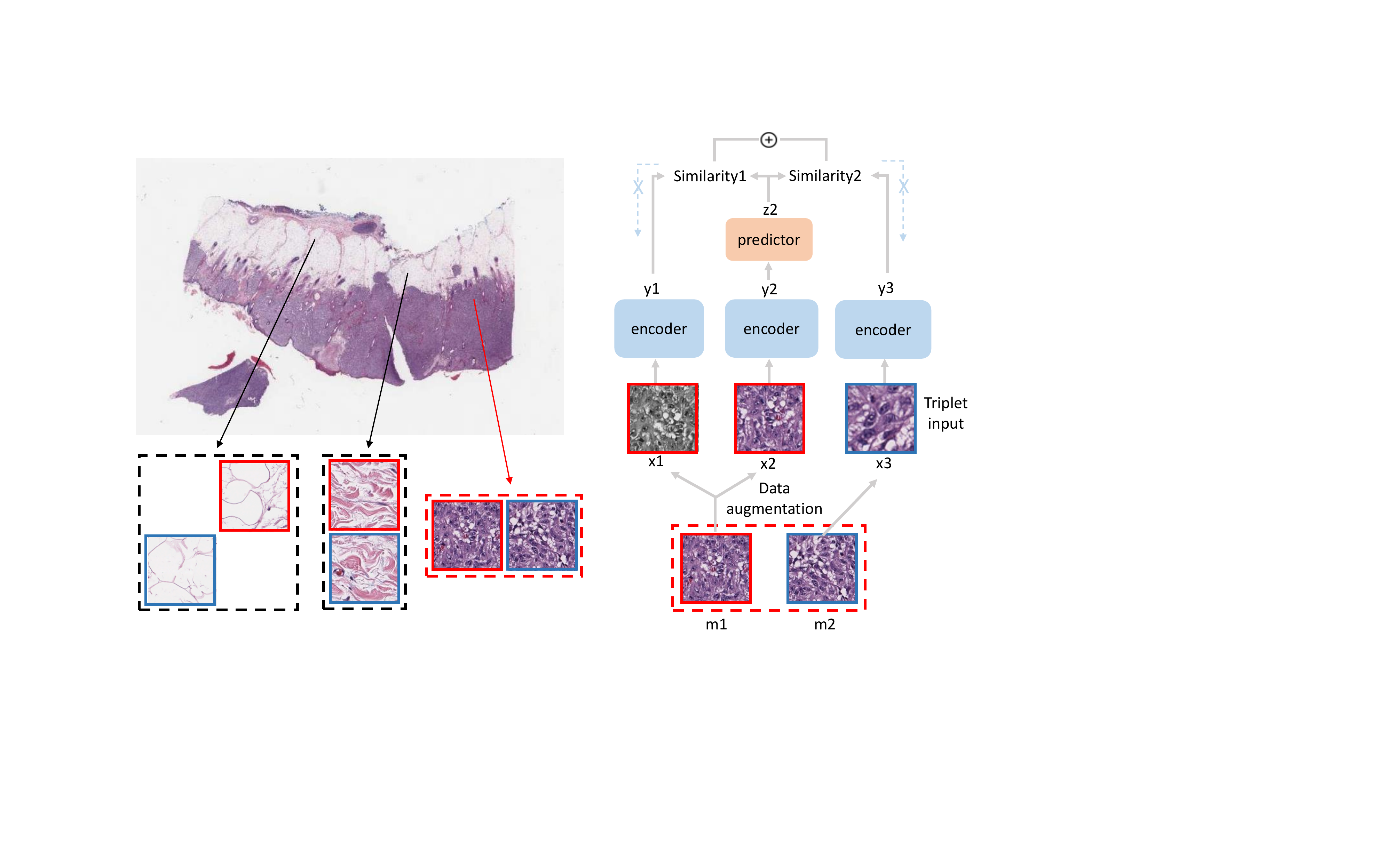}
\end{center}
   \caption{\textbf{Network structure of the proposed SimTriplet}. Adjacent image pairs are sampled from unlabeled pathological images (left panel) for triplet representation learning (right panel). The GigaPixel pathological images provide large-scale "positive pairs" from nearby image patches for SimTriplet. Each triplet consists of two augmentation views from $m_{1}$ and one augmentation view from $m_{2}$. The final loss maximizes both inter-sample and intra-sample similarity as a representation learning.}
\label{fig2:pipeline}
 \end{figure}

In this paper, we propose a simple triplet based representation learning approach (SimTriplet), taking advantage of the multi-view nature of pathological images, with effective learning by using only a single GPU with 16GB memory. We present a triplet similarity loss to maximize the similarity between two augmentation views of same image and between adjacent image patches. The contribution of this paper is three-fold:

$\bullet$ The proposed SimTriplet method takes advantage of the multi-view nature of medical images beyond self-augmentation.

$\bullet$ This method minimizes both intra-sample and inter-sample similarities from positive image pairs, without the needs of negative samples. 

$\bullet$ The proposed method can be trained using a single GPU setting with 16GB memory, with batch size = 128 for 224$\times$224 images, via mixed precision training.

\section{Methods}

\begin{figure}[t]
\begin{center}
\includegraphics[width=0.85\linewidth]{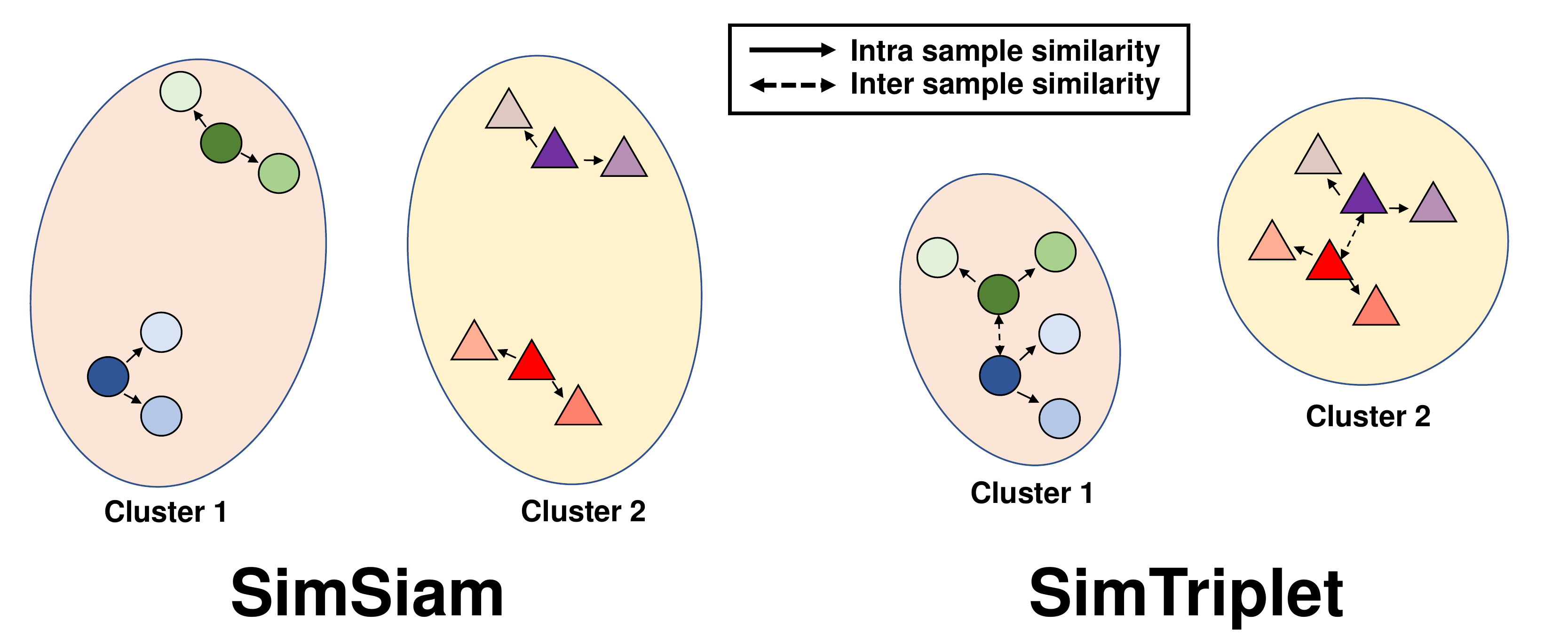}
\end{center}
   \caption{\textbf{Compare SimTriplet with SimSiam}. SimSiam network maximizes intra-sample similarity by minimizing distance between two augmentation views from the same image. The proposed SimTriplet model further enforce the inter-sample similarity from positive sample pairs.}
\label{fig3:simtriplet}
 \end{figure}

The principle network of SimTriplet is presented in Fig~\ref{fig2:pipeline}. The original SimSiam network can be interpreted as an iterative process of two steps: (1) unsupervised clustering and (2) feature updates based on clustering (similar to K-means or EM algorithms)~\cite{chen2020exploring}. By knowing the pairwise information of nearby samples, the SimTriplet aims to further minimize the distance between the "positive pairs" (images from the same classes) in the embedding space (Fig.~\ref{fig3:simtriplet}). 
In the single GPU setting with batch size 128, SimTriplet provides more rich information for the clustering stage. 

\subsection{Multi-view nature of medical images}
In many medical image analysis tasks, multi-view (or called multi-instance) imaging samples from the same patient or the same tissue can provide complementary representation information. For pathological images, the nearby image patches are more likely belong to the same tissue type. Thus, the spatial neighbourhood on a WSI provide rich "positive pairs" (patches with same tissue types) for triplet representation learning. Different from~\cite{hoffer2015deep}, all samples in our triplets are positive samples, inspired by~\cite{chen2020exploring}. To train SimTriplet, we randomly sample image patches as well as their adjacent patches (from one of eight nearby locations randomly) as positive sample pairs from the same tissue type. 

\subsection{Triplet representation learning}
Our SimTriplet network forms a triplet from three randomly augmented views by sampling positive image pairs (Fig.~\ref{fig2:pipeline}). The three augmented views are fed into the encoder network. The encoder network consists of a backbone network (ResNet-50~\cite{he2016deep}) and a three-layer multi-layer perceptron (MLP) projection header. The three forward encoding streams share the same parameters. Next, an MLP predictor is used in the middle path. The predictor processes the encoder output from one image view to match with the encoder output of two other image views. We applies stop-gradient operations to two side paths. When computing loss between predictor output 
and image representation from encoder output, encoded representation is regarded as constant~\cite{chen2020exploring}. Two encoders on side paths will not be updated by back propagation. We used negative cosine similarity Eq.(\ref{eq:cos}) between different augmentation views of (1) the same image patches, and (2) adjacent image patches as our loss function. For example, image $m_{1}$ and image $m_{2}$ are two adjacent patches cropped from the original whole slide image (WSI). $x_{1}$ and $x_{2}$ are randomly augmented views of image $m_{1}$, while $x_{3}$ is the augmented view of image $m_{2}$. Representation $y_{1}$, $y_{2}$ and $y_{3}$ are encoded from augmented views by encoder. $z_{1}$, $z_{2}$ and $z_{3}$ are the representation processed by the predictor. 

\begin{equation}
    \mathcal{C}(p,q)=-\frac{p}{\|p\|}_{2}\cdot\frac{q}{\|q\|}_{2}
    \label{eq:cos}
\end{equation}

$\mathcal{L}_{Intra sample}$ is the loss function to measure the similarities between two augmentation views $x_{1}$ and $x_{2}$ of image $m_{1}$ as seen in Eq.(\ref{eq:sim1}). 
\begin{equation}
    \mathcal{L}_{Intra sample} = \frac{1}{2}\mathcal{C}(y_{1},z_{2}) + \frac{1}{2}\mathcal{C}(y_{2},z_{1})
    \label{eq:sim1}
\end{equation}

$\mathcal{L}_{Inter sample}$ is the loss function to measure the similarities between two augmentation views $x_{2}$ and $x_{3}$ of adjacent image pair $m_{1}$ and $m_{2}$  as in Eq.(\ref{eq:sim2}). 
\begin{equation}
    \mathcal{L}_{Inter sample} = \frac{1}{2}\mathcal{C}(y_{2},z_{3}) + \frac{1}{2}\mathcal{C}(y_{3},z_{2})
    \label{eq:sim2}
\end{equation}
The triplet loss function as used in our SimTriplet network is defined as: 
\begin{equation}
    \mathcal{L}_{total} = \mathcal{L}_{Intra sample} + \mathcal{L}_{Inter sample}
\end{equation}

$\mathcal{L}_{Intra sample}$ minimizes the distance between different augmentations from the same image. $\mathcal{L}_{Inter sample}$ minimizes the difference between nearby image patches.

\subsection{Expand batch size via mix precision training}
Mix precision training~\cite{micikevicius2018mixed} was invented to offer significant computational speedup and less GPU memory consumption by performing operations in half-precision format. The minimal information is stored in single-precision to retain the critical parts of the training. By implementing the mix precision to SimTriplet, we can extend the batch size from 64 to 128 to train images with 224$\times$224 pixels, using a single GPU with 16GB memory. The batch size 128 is regarded as a decent batch size in SimSiam~\cite{chen2020exploring}.


\section{Data and Experiments}
\subsection{Data}
\textbf{Annotated data}. We extracted image patches from seven melanoma skin cancer Whole Slide Images (WSIs) from the Cancer Genome Atlas (TCGA) Datasets (TCGA Research Network: https://www.cancer.gov/tcga). From the seven annotated WSIs, 4698 images from 5 WSIs were obtained for training and validation, while 1,921 images from 2 WSIs were used for testing. Eight tissue types were annotated as : blood vessel (353 train 154 test), epidermis (764 train 429 test), fat (403 train 137 test), immune cell (168 train 112 test), nerve (171 train 0 test), stroma (865 train 265 test), tumor (1,083 train 440 test) and ulceration (341 train 184 test). 

Following~\cite{raju2020graph,zhao2020predicting}), each image was a 512$\times$512 patch extracted from 40$\times$ magnification of a WSI with original pixel resolution 0.25 micron meter. The cropped image samples were annotated by a board-certified dermatologist and confirmed by another pathologist. Then, the image patches were resized to 128$\times$128 pixels. Note that the 224$\times$224 image resolution provided 1.8$\%$ higher balance accuracy (based on our experiments) using the supervised learning. We chose 128$\times$128 resolution for all experiments for a faster training speed.

\textbf{Unlabeled data}. Beyond the 7 annotated WSIs, additional 79 WSIs without annotations were used for training contrastive learning models. The 79 WSIs were all available and usable melanoma cases from TCGA. The number and size of image patches used for different contrastive learning strategies are described in $\mathsection$\textbf{Experiment}.


\subsection{Supervised learning}

We used ResNet-50 as the backbone in supervised training, where the optimizer is Stochastic Gradient Descent (SGD)~\cite{bottou2010large} with the base learning rate $lr=0.05$. The optimizer learning rate followed (linear scaling~\cite{goyal2017accurate}) $lr\times$BatchSize$/256$. We used 5-fold cross validation by using images from four WSIs for training and image from the remaining WSI for validation. We trained 100 epochs and selected best model based on validation. When applying the trained model on testing images, the predicted probabilities from five models were averaged. Then, the class with the largest ensemble probability was used as the predicted label.

\subsection{Training contrastive learning benchmarks}
We used the SimSiam network~\cite{chen2020simple} as the baseline method of contrastive learning. Two random augmentations from the same image were used as training data. In all of our self-supervised pre-training, images for model training were resized to $128\times128$ pixels. We used momentum SGD as the optimizer. Weight decay was set to 0.0001. Base learning rate was $lr=0.05$ and batch size equals 128. Learning rate was $lr\times$BatchSize$/256$, which followed a cosine decay schedule~\cite{loshchilov2017sgdr}. Experiments were achieved only on a single GPU with 16GB memory. Models were pre-trained for $39,500/128\times400\approx127,438$ iterations. 79 unlabeled WSIs were used for self-supervised pre-training. We randomly cropped 500 images from each WSI and resized them to $128\times128$ pixels. 39,500 images in total serve as the original training data. 

Following MICLe~\cite{azizi2021big}, we employed multi-view images as two inputs of the network. Since we did not use negative samples, multi-view images was trained by SimSiam network instead of SimCLR. For each image in the original training dataset, we cropped one patch which is randomly selected from its eight adjacent patches consisting of an adjacent images pairs. We had 79,000 images (39,500 adjacent pairs) as training data. Different from original SimSiam, network inputs were augmentation views of an adjacent pair. Referring to \cite{chen2020exploring}, we applied our data on SimSiam network. First, we used 39,500 images in original training dataset to pre-train on SimSiam. To see the impact of training dataset size, we randomly cropped another 39,500 images from 79 WSIs for training on a larger dataset of 79,000 images. We then used training data from the MICLe experiment to train the SimSiam network.

\subsection{Training the proposed SimTriplet}
The same 79,000 images (39,500 adjacent pairs) were used to train the SimTriplet. Three augmentation views from each adjacent pair were used as network inputs. Two augmentation views were from one image, while the other augmentation view was augmented from adjacent images. Three augmentation views were generated randomly, where the augmentation settings were similar with the experiment on SimSiam~\cite{chen2020simple}. Batch size was 128 and experiment run on a single 16GB memory GPU.

\begin{figure}[t]
\begin{center}
\includegraphics[width=1\linewidth]{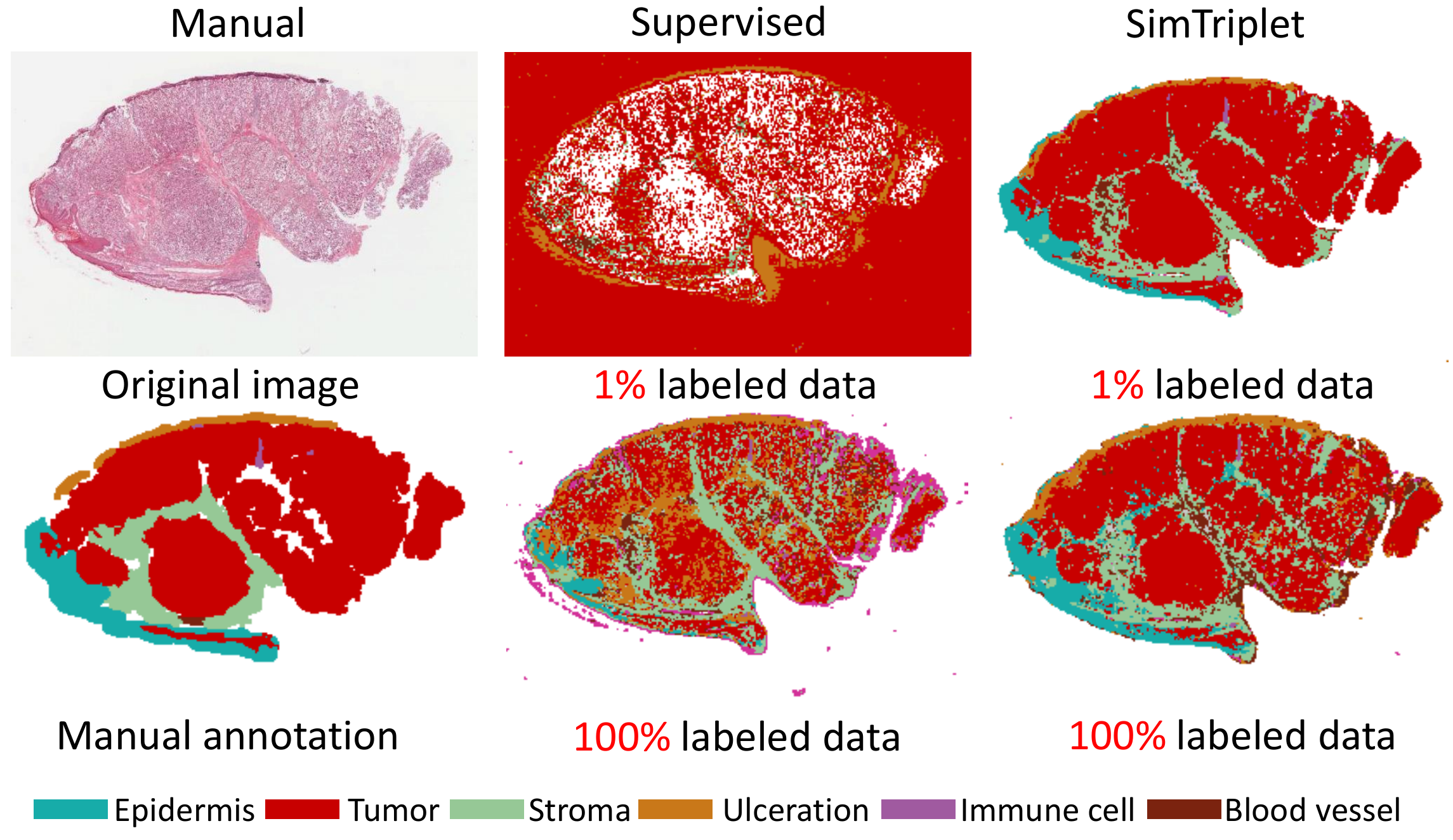}
\end{center}
   \caption{\textbf{Visualization of classification results.} One tissue sample is manually segmented by our dermatologist (via QuPath software) to visually compare the classification results. The contrasting learning achieved superior performance compared with supervised learning, even using only 1$\%$ of all available labeled data.}
\label{fig4:seg}
 \end{figure}

\subsection{Linear evaluation (fine tuning)}
To apply the self-supervised pre-training networks, as a common practice, we froze the pretrained ResNet-50 model by adding one extra linear layer which followed the global average pooling layer. When finetuning with the annotated data, only the extra linear layer was trained. We used the SGD optimizer to train linear classifier with a based (initial) learning rate $lr$=30, weight decay=0, momentum=0.9, and batch size=64 (follows~\cite{chen2020exploring}). 
The same annotated dataset were used to finetune the contrastive learning models as well as to train supervised learning. Briefly, 4,968 images from 5 annotated WSIs were divided into 5 folders. We used 5-fold cross validation: using four of five folders as training data and the other folder as validation. We trained linear classifiers for 30 epochs and selected the best model based on the validation set. The pretrained models were applied to the testing dataset (1,921 images from two WSIs). As a multi-class setting, macro-level average F1 score was used~\cite{attia2018multilingual}. Balanced accuracy was also broadly used to show the model performance on unbalanced data~\cite{5597285}. 

\section{Results}

\subsubsection{Model classification performance.}
F1 score and balanced accuracy were used to evaluate different methods as described above. We trained a supervised learning models as the baseline. From Table~\ref{table:performance}, our proposed SimTriplet network achieved the best performance compared with the supervised model and SimSiam network~\cite{chen2020exploring} with same number of iterations. To show a qualitative result, a segmentation of a WSI from test dataset is shown in Fig.~\ref{fig4:seg}.  

\begin{table}
\caption{Classification performance.}
\centering
\begin{tabular}{p{2.5cm}p{1.6cm}p{1.2cm}p{1.5cm}p{1.5cm}}
\toprule
 Methods & Unlabeled Images & Paired Inputs & F1 \quad \quad Score &Balanced Acc \\
\midrule
Supervised  &0  &  & 0.5146 & 0.6113  \\ 
\midrule
MICLe~\cite{azizi2021big}*  & 79k & \checkmark & 0.5856 & 0.6666  \\ 
SimSiam~\cite{chen2020exploring}  & 39.5k & & 0.5421 & 0.5735  \\
SimSiam~\cite{chen2020exploring}  & 79k &\checkmark  & 0.6267 & 0.6988  \\ 
SimSiam~\cite{chen2020exploring}  & 79k & & 0.6275 & 0.6958  \\ 
\midrule
SimTriplet (ours)  & 79k & \checkmark & \textbf{0.6477} & \textbf{0.7171} \\ 
\bottomrule
\multicolumn{5}{l}{* We replace SimCLR with SimSiam.}
\label{table:performance}
\end{tabular}
\end{table}

\subsubsection{Model performance on partial training data.}
To evaluate the impact of training data number, we trained a supervised model and fine-tuned a classifier of the contrastive learning model on different percentages of annotated training data (Table~\ref{table:ratio}). Note that for 1$\%$ to 25$\%$, we ensure different classes contribute a similar numbers images to address the issue that the annotation is highly imbalanced.  

\begin{table}
\caption{Balanced Acc of using different percentage of annotated data.}
\centering
\begin{tabular}{ccccc}
\toprule
$\quad\quad$  Methods$\quad\quad$  & \multicolumn{4}{c}{Percentage of Used Annotated Training Data} \\
\midrule
$\quad$  &  $\quad$ $1\%$ $\quad$ &$\quad$ $10\%$ $\quad$& $\quad$$25\%$ $\quad$& $\quad$$100\%$ $\quad$\\
\midrule
Supervised  & 0.0614 & 0.3561 & 0.4895 & 0.6113 \\ 
SimSiam~\cite{chen2020exploring}  & 0.7085 & 0.6864 & 0.6986 &  0.6958\\ 
SimTriplet  & \textbf{0.7090} & \textbf{0.7110} & \textbf{0.7280} & \textbf{0.7171} \\ 
\bottomrule
\label{table:ratio}
\end{tabular}
\end{table}

\section{Conclusion}
In this paper, we proposed a simple contrastive representation learning approach, named SimTriplet, advanced by the multi-view nature of medical images. Our proposed contrastive learning methods maximize the similarity between both self augmentation views and pairwise image views from triplets. Moreover, our model can be efficiently trained on a single GPU with 16 GB memory. The performance of different learning schemes are evaluated on WSIs, with large-scale unlabeled samples. The proposed SimTriplet achieved superior performance compared with benchmarks, including supervised learning baseline and SimSiam method. The contrastive learning strategies showed strong generalizability by achieving decent performance by only using 1$\%$ labeled data.




%
%
\bibliographystyle{splncs04}
\bibliography{main}
%




\end{document}